\pdfoutput=1

\documentclass[11pt]{article}

\usepackage{acl}

\usepackage{times}
\usepackage{latexsym}
\usepackage{caption}
\usepackage{graphicx}
\usepackage{subcaption}
\usepackage{csquotes}
\usepackage{amsmath, mathtools, multirow}
\usepackage{url}
\usepackage{tabularx}
\usepackage{csquotes}

\usepackage[T1]{fontenc}

\usepackage[utf8]{inputenc}

\usepackage{microtype}

\newcommand{\mg}[1]{{\textcolor{black}{#1}}{\bf}}
\newcommand{\sara}[1]{{\textcolor{black}{#1}}{\bf}}

\title{Length Adaptive Average Lagging}
\title{Over-Generation Cannot Be Rewarded:\\ Length-Adaptive Average Lagging for Simultaneous Speech Translation}

\author{Sara Papi\textsuperscript{$\diamondsuit,\triangle$}, Marco Gaido\textsuperscript{$\diamondsuit,\triangle$}, Matteo Negri\textsuperscript{$\diamondsuit$}, Marco Turchi\textsuperscript{$\diamondsuit$} \\
  \textsuperscript{$\diamondsuit$}Fondazione Bruno Kessler \\
  \textsuperscript{$\triangle$}University of Trento \\
  \texttt{\{spapi, mgaido, negri, turchi\}@fbk.eu}}

\begin{document}
\maketitle
\begin{abstract}

Simultaneous speech translation (SimulST) systems aim at generating their output with the lowest possible latency, which is normally computed in terms of Average Lagging (AL). In this paper we highlight that, despite its widespread adoption, AL provides underestimated scores for systems that generate longer predictions compared to the corresponding references. We also show that this problem has practical relevance, as recent SimulST systems have indeed a tendency to over-generate. As a solution, we propose LAAL (Length-Adaptive Average Lagging), a modified version of the metric that takes into account the over-generation phenomenon and allows for unbiased evaluation of both under-/over-generating systems.

\end{abstract}

\section{Introduction}
Simultaneous
speech-to-text translation (SimulST) is the task in which the generation of the textual translation in the target language starts before the entire audio input in the source language has been ingested by the model. The need to have high quality translations in the shortest possible time 
therefore becomes the main objective of SimulST systems,
which 
have to satisfy 
specific
time requirements 
depending
on the 
application scenarios. 
These 
requirements are usually expressed in terms of 
latency, that is the elapsed time
between the pronunciation of a word and the generation of its textual translation. 
How latency is measured hence plays a crucial role in 
systems
evaluation.

The SimulST task was initially addressed using cascaded models \citep{10.2307/30219116,oda-etal-2014-optimizing} that divide the translation process into two steps: simultaneous automatic speech recognition \citep{NIPS2016_312351bf,rao2017exploring}, and simultaneous machine translation  \citep{cho2016neural,gu-etal-2017-learning}. 
For this reason, the first latency metrics were 
designed to evaluate simultaneous machine translation (SimulMT) systems \citep{cho2016neural,cherry2019thinking,elbayad-etal-2020-online}.
Among them, Average Lagging -- AL -- \citep{ma-etal-2019-stacl} is the most popular one, and its adaptation to SimulST by \citet{ma-etal-2020-simuleval} has become a 
widely adopted \citep{ma-etal-2020-simulmt,zeng-etal-2021-realtrans,chen-etal-2021-direct,liu-etal-2021-cross}
\textit{de facto} standard.\footnote{For instance, the IWSLT SimulST Shared Task \citep{anastasopoulos-etal-2021-findings} 
relies on
AL to divide the systems in different latency regimes (low, medium, high) and BLEU \citep{post-2018-call} to rank the them based on translation quality.}
The adaptation by \citet{ma-etal-2020-simuleval} 
sparks from a weakness observed in the original formulation of the 
metric. Being
susceptible to \textit{under-generation}, it results in biased evaluations favouring systems that produce shorter predictions compared to the reference. However, though successful in correcting this behaviour, the proposed adaptation did not consider the opposite case of \textit{over-generation}, which occurs when the prediction is longer than the reference.
%

To fill this gap, in this paper we introduce LAAL (Length-Adaptive Average Lagging):\footnote{The code is available at: \url{https://github.com/hlt-mt/FBK-fairseq}.} a simple yet effective extension of AL that takes into account also over-generation and allows for fair SimulST systems comparisons.
%
%
%
%
%
%
%
After 
a brief 
explanation of
AL calculation (Section \ref{sec:AL}), we expose its incorrect behaviour in presence of over-generation phenomena (Section \ref{sec:problems}) and show that over-generation is actually present in the output of recent SimulST systems
(Section \ref{sec:frequency}). Then,  we present 
the new LAAL metric
(Section \ref{sec:LAAL}), whose computation is adjusted at sentence level by looking at the length of  model predictions. 
Through examples, we show that, unlike the previous AL formulation, our metric is able to fairly evaluate both under- and 
over-generating
systems. 
We conclude our work with a discussion (Section \ref{sec:discussion}) about problems that still need to be solved for latency computation, remarking that our proposal 
represents a
first step toward a more reliable assessment of SimulST systems performance.

\section{Average Lagging}
\label{sec:AL}
The idea behind the AL metric is to 
quantify how much time the 
system
is out of sync with the speaker.
In SimulST, the input sequence is represented as a stream of audio speech in the source language $\mathbf{X} = [x_{1}, ..., x_{|\mathbf{X}|}]$ where each element $x_{j}$ is a raw audio segment of duration $T_{j}$, the reference
as a stream of words in the target language $\mathbf{Y^{*}} = [y^{*}_{1}, ..., y^{*}_{|\mathbf{Y^{*}}|}]$, and the model translation
as a stream of predicted words $\mathbf{Y} = [y_{1}, ..., y_{|\mathbf{Y}|}]$. In the simultaneous 
setting,
a system starts to generate a partial hypothesis while it continues to receive an incremental stream of input. 
This implies that, 
to generate
the $y_{i}$ target word at time $j$, it has access to $\mathbf{X}_{1:j} = [x_{1}, ..., x_{j}]$ with $j<|\mathbf{X}|$.

Therefore, the delay with which the $y_{i}$ word is emitted is $d_i = \sum_{i=1}^{j} T_i$.
Using this notation,
in \citep{ma-etal-2020-simuleval}
Average Lagging was initially
defined as follows:
\begin{gather} 
\label{equation:AL}
AL = \frac{1}{\tau'(|\mathbf{X}|)}\sum_{i=1}^{\tau'(|\mathbf{X}|)}d_i - d^*_i \\
\label{equation:di}
d^*_i = (i-1) \cdot \frac{\sum_{j=1}^{|\mathbf{X}|} T_j}{|\textbf{Y}|}
\end{gather}
where $\tau'(|\mathbf{X}|) =\min\{i|d_i = \sum_{j=1}^{|\mathbf{X}|} T_j\}$ 
is
the index of the target token when
the end of the source sentence is reached
and $d^*_i$ represents 
an oracle
that, perfectly in sync with the speaker,
starts to emit words as soon as the speech starts.

However, the authors noticed that this adaptation was not robust for models that tend to stop 
generating the hypothesis
too early, that is systems that under-generate.
This phenomenon is more likely to happen in SimulST than in SimulMT, for which 
AL was first  
proposed. For instance, the presence of long pauses in the speech may induce systems to generate the end of sentence token too early, even if the source utterance is not yet complete.
As observed by the authors,
when this phenomenon occurs, the lagging behind the oracle becomes negative. 
It follows that relatively good latency-quality trade-offs can be achieved thanks to inappropriate AL discounts in case of under-generation, while this does not reflect the reality.
Thus, 
in \citep{ma-etal-2020-simuleval}, Equation \ref{equation:AL} was redefined as:
\begin{equation}
\label{equation:new-di}
    d^*_i = (i-1) \cdot \frac{\sum_{j=1}^{|\mathbf{X}|} T_j}{|\textbf{Y}^*|}
\end{equation}
assuming that the oracle delays $d_i^*$ are computed based on the reference length rather than on the system hypothesis length.



\begin{figure*}[t]
    \centering
    \includegraphics[scale=0.495]{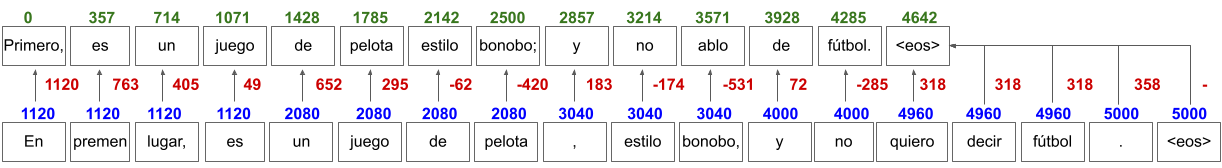}
    \caption{
    Example of AL computation between the oracle delays (in green) and the system delays (in blue). The lagging values (in red) are computed as the difference between the system and the oracle delays (the mapping is represented by an arrow). The $198$ms AL is obtained by dividing the sum of the lagging ($3379$ms) by its count ($17$).
    }
    \label{fig:al_issue}
\end{figure*}

\section{The Problem of Over-Generation}
\label{sec:problems}

In this paper, we point out a major issue of AL that arises in
presence of over-generation. As we will see, 
\mg{AL} improperly favors
\mg{over-generating systems,}
potentially 
\mg{leading} the community to wrong conclusions due to biased evaluations. 
To illustrate how over-generation affects
\mg{AL} 
\mg{computation,}
we consider a real example
from the English$\rightarrow$Spanish (en-es) section of MuST-C \citep{CATTONI2021101155} tst-COMMON translated by
the state-of-the-art 
Cross Attention Augmented Transducer (CAAT)
system \mg{\citep{liu-etal-2021-cross}.}

As  shown in
Figure \ref{fig:al_issue},
the  prediction suffers from over-generation, especially in the first part of the sentence where more target words 
are produced compared to the reference.
The system translates \enquote{\textit{En primer lugar,}} instead of \enquote{\textit{Primero,}}, 
forcing
all the predicted words
to compare with the successive word 
in
the reference. For instance, \enquote{\textit{es}} in the CAAT output is computed against \enquote{\textit{juego}} in the oracle and its very low lagging ($49$ms) is a considerable underestimation of the correct lagging with respect to the \enquote{\textit{es}} word in the oracle ($763$ms).
Likewise, \enquote{\textit{de}} is assigned a negative lagging ($-62$ms) instead of the $652$ms 
delay with respect to the
time of \enquote{\textit{de}} in the oracle.
Finally, all the words generated before the end of the utterance (in our example $5000$ms) and exceeding the reference length are compared with the last word of the oracle; the same happens to the first word emitted when the utterance is over, while the other words after the end of the utterance are ignored.
As a result, the AL of CAAT output for this sentence is $198$ms,
an extremely low latency that does not reflect the truth.
Indeed, if we correctly align and compare the words in the system output and the oracle, we see that lagging is on average $846$ms.\footnote{The detailed calculation 
is
presented in Appendix \ref{sec:calculations}.}
This represents a problem, since the AL metric is rewarding an
over-generating
system, 
potentially hindering a fair comparison with other models. 

In light of these observations, two questions arise. First, \textit{what are the conditions leading to biased, i.e. underestimated, AL values?}
The example above shows that the problem arises
when: \textit{i)} the system over-generates,
and \textit{ii)} 
the over-generated words appear before the utterance ends (the earlier in time, the lower the final AL score will be).
Second, \textit{why was this problem overlooked when AL was introduced?}
To answer this question, recall
that 
AL for SimulST was initially proposed to evaluate 
systems 
showing a biased behaviour toward under-generation, 
with predictions length reaching at most the reference length.
Indeed,
earlier SimulST systems \citep{ma-etal-2020-simulmt} 
were
designed to emit only one word 
at
each time step.
Consequently, even 
in the 
case of over-generation,
it was extremely unlikely for the number of words emitted before the end of the utterance to be higher than the number of words in the 
reference. Moreover, additional words (if any) were generated only once the utterance was over.
As mentioned above, 
the current
AL implementation ignores all but the first word emitted at the end of the 
utterance. Therefore,
the over-generation occurring at the end of the utterance
does not affect the metric computation.
\sara{Instead,}
AL is not robust to over-generation if it occurs before 
the end of the utterance, an extremely unlikely behavior in early systems but frequent in more recent ones, as we will see in the next section.

\section{Over-generation frequency}
\label{sec:frequency}
To quantify the impact of over-generation on system evaluation,
we 
check
how 
frequently it occurs in the output of three
SimulST 
systems: CAAT, the wait-k model by \citet{ma-etal-2020-simulmt}, and an offline model with the wait-k policy applied only at inference time \citep{papi2022does,gaido-etal-2022-efficient}.
%
%
%
We run three systems on
the 
en-es section of MuST-C tst-COMMON by varying the $k$ value at inference time in the range 
$\{3, 5, 7, 9, 11\}$.
%
%
%
%
%
We measure over-generation in terms of average word length difference (AWLD)
between 
systems predictions and the corresponding 
references, that is:
\begin{equation}
\label{equation:AWLD}
    \text{AWLD} = \frac{1}{N} \sum_{s=1}^{N} |\mathbf{Y}| - |\mathbf{Y}^*|
\end{equation}
where $N$ is the number of samples in the corpus. 
Accordingly, 
positive AWLD values indicate that system predictions are on average longer than the reference 
(over-generation), while negative
values indicate systems
tendency to under-generate.



\begin{table}[hbt]
    \centering
    \small
    \setlength{\tabcolsep}{4pt}
    \begin{tabular}{c|c|c|c|c|c}
        \hline
        \texttt{Model} & \texttt{k=3} & \texttt{k=5} & \texttt{k=7} & \texttt{k=9} & \texttt{k=11} \\
         \hline
         wait-k & -5.57 & -3.82 & -2.30 & -1.13 & -0.74 \\
         offline wait-k & 0.48 & 0.49 & 0.53 & 0.74 & 0.80 \\
         CAAT & 1.57 & 0.96 & 0.61 & 0.35 & 0.18 \\
        \hline
    \end{tabular}
    \caption{AWLD on MuST-C en-es tst-COMMON.}
    \label{tab:word_diff}
\end{table}

Table \ref{tab:word_diff} 
shows
that the wait-k system under-generates -- as already noticed by \citet{ma-etal-2020-simulmt} -- while 
both CAAT and offline wait-k ones over-generate.
In addition, while for the offline wait-k model the over-generation phenomenon is quite constant 
for each $k$ value, for CAAT 
this
diminishes
 as 
 $k$  increases.
This indicates
that over-generation is not an isolated phenomenon 
affecting
only
few sentences. 
%
On the contrary,
it frequently occurs 
and
automatic evaluation should take
this
into account.

\section{Length Adaptive Average Lagging}
\label{sec:LAAL}
Based on the observations made in Sections \ref{sec:problems} and \ref{sec:frequency}, we propose 
LAAL (Length-Adaptive Average Lagging), a 
modified version of AL accounting also for the over-generation phenomena.
LAAL defines the oracle delays by dividing the utterance length by the maximum between the reference and the model prediction length.
Specifically, we consider the reference length when the prediction is shorter (under-generation), and the prediction length when the prediction is longer (over-generation).
This means that the correction is made at sentence level, 
making the metric applicable to a system disregarding its under- or over-generation tendency. 

\begin{figure}[tb]
\centering
\includegraphics[width=0.45\textwidth]{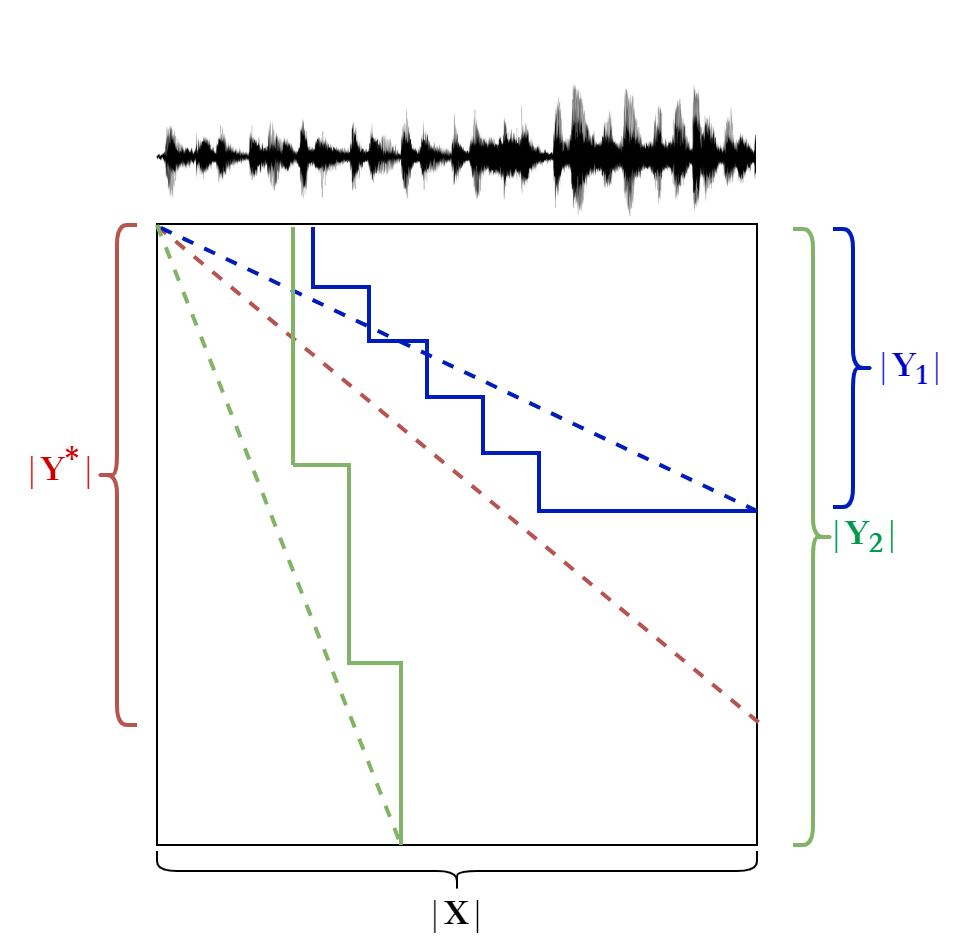}
\caption{Example of under-generation (in blue), and over-generation (in green). The reference translation is represented by the red dashed line.}
\label{fig:ideal}
\end{figure}

Figure \ref{fig:ideal} shows both the 
under-generation
(in blue) and the 
over-generation
(in green) cases.
Analyzing the 
under-generation
case, we can clearly see the motivation behind the correction made by \citet{ma-etal-2020-simuleval}: if we consider the prediction length ($|\mathbf{Y_1}|$) in the AL computation and the prediction is too short, the system is favoured
since negative delays are summed (the blue straight line is mainly below the blue dashed line). 
A more reliable evaluation is obtained by considering the reference length ($|\mathbf{Y^*}|$), since the straight blue line is always above the dashed red line.
By analyzing the over-generation case, we observe the opposite problem: if we consider the reference length ($|\mathbf{Y^*}|$) in the AL computation and the prediction is too long ($|\mathbf{Y_2}|>|\mathbf{Y^*}|$), the system is favoured since negative delays are summed (the straight green line stays almost always below the red dashed line).
This can be corrected by considering the prediction length ($|\mathbf{Y_2}|$) instead. 
%
%
%
%
%
%
Therefore, to make a more reliable 
evaluation where neither under-generation nor over-generation are rewarded,
we have to take the maximum between $|\mathbf{Y^*}|$ and $|\mathbf{Y}|$ in the delay computation (the two conditions are: $|\mathbf{Y^*}|$ if $|\mathbf{Y}|\leq|\mathbf{Y^*}|$, and  $|\mathbf{Y}|$ if $|\mathbf{Y}|>|\mathbf{Y^*}|$).
Accordingly, 
Equation \ref{equation:new-di} can be modified to obtain LAAL as:
\begin{equation}
\label{equation:ALO}
    d^*_i = (i-1) \cdot \frac{\sum_{j=1}^{|\mathbf{X}|} T_j}{\max\{|\mathbf{Y}|,|\mathbf{Y^{*}|\}}}
\end{equation}

The difference between 
applying AL and LAAL to 
evaluate
our three systems
is shown
in Table \ref{tab:LAAL}. As we can see, the LAAL 
of the wait-k system
is
almost equal to the AL,
with differences from $17$ to $73$ms.
Conversely, 
for the
offline wait-k system we notice a quite constant increment in LAAL of about $120$ms
while for the
CAAT system
we observe that LAAL is visibly greater than AL, with differences from $117$ to $283$ms that are more marked at low latency. These differences are 
coherent with the over-generation trend 
observed in
Table \ref{tab:word_diff}.

\begin{table}[hbt]
    \centering
    \small
    \setlength{\tabcolsep}{4pt}
    \begin{tabular}{c|c|c|c|c|c|c}
        \hline
        \texttt{Model} & \texttt{Metric} & \texttt{k=3} & \texttt{k=5} & \texttt{k=7} & \texttt{k=9} & \texttt{k=11} \\
         \hline
        \multirow{2}{*}{wait-k} & AL & 1761 & 1970 & 2272 & 2582 & 2931\\
         & LAAL & 1778 & 2001 & 2332 & 2655 & 3003 \\
         \hline
        offline & AL & 1522 & 1959 & 2463 & 2926 & 3350  \\
         wait-k & LAAL & 1682 & 2093 & 2588 & 3043 & 3457 \\
         \hline
         \multirow{2}{*}{CAAT} & AL & 735 & 1149 & 1533 & 1905 & 2265 \\
         & LAAL & 1018 & 1365 & 1708 & 2046 & 2382 \\
        \hline
    \end{tabular}
    \caption{AL and LAAL results in ms of the wait-k and CAAT systems on MuST-C en-es tst-COMMON.}
    \label{tab:LAAL}
\end{table}

%
Going back to the example in Figure \ref{fig:al_issue},  the latency value computed with LAAL is $707$ms. Compared to the AL value of $198$ms, this is much closer to the real measure of 
$846$ms
calculated in Section \ref{sec:problems}. In light of these observations, we can conclude that the LAAL metric gives a more reliable evaluation of the SimulST systems compared to AL.


\section{Limitations} 
\label{sec:discussion}
The proposed LAAL metric is a
first step toward a more accurate evaluation of 
SimulST systems.
Although in this work we focused on the over-generation problem, we 
did not address
another limitation of AL (and, in turn, of LAAL). The problem
is
that, as shown in Section \ref{sec:AL}, AL compares the system output with an oracle that emits only one word at each time step, each one with a fixed word duration.\footnote{In our
example in Figure \ref{fig:al_issue}, the word duration is $357$ms and is computed
dividing the source audio duration ($5000$ms) by the reference length ($14$).} 
This means that, in its computation, we assume that the reference words are uniformly distributed in each utterance. However, considering that the amount of information contained in audio segments of the same length could be extremely different, this represents an unrealistic approximation. For instance, a
speech segment can contain silences, long pauses, and
the speech rate can vary considerably. As a consequence, the latency scores obtained can still largely differ from the latency experienced by the user. This advocates for the development of more human-centric solutions that go beyond AL-like metrics despite their success, accounting for different audio phenomena and their impact on the actual latency perceived by the users, also considering the visualization strategy selected \citep{karakanta2021simultaneous,papi2021visualization}. We leave this line of investigation for future work.

\section{Conclusions}
We
showed through examples based on real 
systems that the current Average Lagging
computation is inadequate to correctly 
measure SimulST performance
in presence of over-generation phenomena. To overcome this problem, we proposed 
Length-Adaptive Average Lagging (LAAL), a latency metric that can 
effectively handle both under- and over-generation at sentence 
level, leading to a more reliable  evaluation of SimulST systems.

\section*{Acknowledgments}
This work has been carried out as part of the project Smarter Interpreting (\url{https://kunveno.digital/}) financed by CDTI Neotec funds.



\bibliography{anthology,custom}
\bibliographystyle{acl_natbib}

\appendix
\section{Example Manual Latency Calculation}
\label{sec:calculations}
To manually compute the latency measure of the example shown in Figure \ref{fig:al_issue}, we compare the model output words to the reference words of the oracle by correctly aligning them. For instance \enquote{\emph{En premen lugar,}} of the model prediction is aligned to \enquote{\emph{Primero,}} of the oracle, \enquote{\emph{es}} of the model prediction to \enquote{\emph{es}} of the oracle, and so on.
Therefore, the lagging calculation will be the following:
\begin{equation*}
\small
\begin{split}
 & (1120 - 0) + (1120 - 357) + (2080 - 714) + \\
 & (2080 - 1071) + (2080 - 1428) + (2080 - 1785) + \\
 & (3040 - 2142) + (3040 - 2500) + (4000 - 2857) + \\
 & (4000 - 3214) + (4960 - 3571) + (4960 - 4285) + \\
 & (5000 - 4642) = 10994
\end{split}
\end{equation*}
Then, we divide the lagging sum of $10994$ms by their count ($13$) to obtain the latency of $846$ms.

\end{document}